\documentclass[conference]{IEEEtran}
\IEEEoverridecommandlockouts
\usepackage{cite}
\usepackage{amsmath,amssymb,amsfonts}
\usepackage{algorithmic}
\usepackage{graphicx}
\usepackage{subfig}
\usepackage{textcomp}
\usepackage{xcolor}
\usepackage{colortbl}
\usepackage{nicematrix}
\usepackage{bbold}
\usepackage{bm}
\usepackage{float}
\def\BibTeX{{\rm B\kern-.05em{\sc i\kern-.025em b}\kern-.08em
    T\kern-.1667em\lower.7ex\hbox{E}\kern-.125emX}}
\begin{document}

\title{Constrained Recurrent Bayesian Forecasting for Crack Propagation\\
}

\makeatletter
\newcommand{\linebreakand}{%
  \end{@IEEEauthorhalign}
  \hfill\mbox{}\par
  \mbox{}\hfill\begin{@IEEEauthorhalign}
}
\makeatother
\author{\IEEEauthorblockN{Sara Yasmine OUERK}
\IEEEauthorblockA{\textit{IRT SystemX}\\
Palaiseau, France \\
sara-yasmine.ouerk@irt-systemx.fr}
\and
\IEEEauthorblockN{Olivier VO VAN}
\IEEEauthorblockA{\textit{SNCF}\\
Saint-Denis, France \\
olivier.vovan@sncf.fr}
\and 
\IEEEauthorblockN{Mouadh YAGOUBI}
\IEEEauthorblockA{\textit{IRT SystemX}\\
Palaiseau, France \\
mouadh.yagoubi@irt-systemx.fr}

}

\maketitle

\begin{abstract}
Predictive maintenance of railway infrastructure, especially railroads, is essential to ensure safety. However, accurate prediction of crack evolution represents a major challenge due to the complex interactions between intrinsic and external factors, as well as measurement uncertainties. Effective modeling requires a multidimensional approach and a comprehensive understanding of these dynamics and uncertainties. Motivated by an industrial use case based on collected real data containing measured crack lengths, this paper introduces a robust Bayesian multi-horizon approach for predicting the temporal evolution of crack lengths on rails. This model captures the intricate interplay between various factors influencing crack growth. Additionally, the Bayesian approach quantifies both epistemic and aleatoric uncertainties, providing a confidence interval around predictions. To enhance the model's reliability for railroad maintenance, specific constraints are incorporated. These constraints limit non-physical crack propagation behavior and prioritize safety. The findings reveal a trade-off between prediction accuracy and constraint compliance, highlighting the nuanced decision-making process in model training. This study offers insights into advanced predictive modeling for dynamic temporal forecasting, particularly in railway maintenance, with potential applications in other domains.
\end{abstract}


\section{Introduction}
The rail network is a crucial structure for rail transport, ensuring the connectivity and mobility of millions of passengers and goods every day. Predictive maintenance of rail infrastructure, particularly railroads, is essential for guaranteeing the system's safety and longevity. However, predicting the temporal evolution of cracks in the network remains a complex challenge that requires an innovative, multi-dimensional approach.\\
Crack propagation in rails depends not only on intrinsic factors, linked with the production process and the chemical composition, but also on a range of external features, such as weather conditions, load from rolling stock, and other operational parameters. Modeling these multivariate time series requires an in-depth understanding of the intricate dynamics underlying the interaction between these factors.\\
This paper introduces a robust Bayesian multi-horizon model for forecasting the temporal progression of rail crack lengths. The purpose of this model is to capture the complex interactions between various external factors influencing crack growth, employing recurrent Bayesian methods. Additionally, the Bayesian approach aims to quantify two types of uncertainty: epistemic and aleatoric, thus providing a confidence interval around its predictions. To enhance the predictive efficiency of this model, it is crucial to integrate important constraints related to railway maintenance management. These constraints are not only relevant to the railway industry but also offer versatility for potential applications in other domains.
The first constraint introduced prevents the non-physical behavior of crack length reduction over time. This constraint is adaptable to various production systems that require incremental predictions.\\
The second constraint ensures that the model does not underestimate the length of cracks, particularly when they become large. It is referred to as the asymmetry constraint and represents an essential addition, particularly in the critical infrastructure domain. Its implications extend beyond railway maintenance, providing applicability for a wide range of systems where early detection and anticipatory action are essential.\\
The reminder of this paper is organized as follows: Section \ref{sec:related} introduces some related works while section \ref{sec:methodo} is devoted to present the proposed methodology. Section \ref{sec:exp} will describe the experiments that were set up to assess the proposed model. Section \ref{sec:conclusion} summarizes the contribution of this work and suggests directions for future research.

\section{Related Work}
\label{sec:related}
\subsection{Rail Crack Propagation Forecasting}
Rail degradation can be divided into two distinct events: crack initiation and crack propagation. Both depend mainly on same phenomena such as train load and train speed, but follow different physical laws. While Wang et. al  \cite{wang2022} chose a data-based modeling approach that includes both phenomena and directly analyzes and predicts breaks, Ghofrani et. al \cite{ghofrani2021} separated it into two different steps and added a physically informed model to improve predictions. Since maintenance actions differ significantly depending on whether a crack has just initiated or has already propagated, and given that rail breaks tend to be uncommon with constant maintenance and monitoring improvements, it is more effective to analyze propagation separately.\\
The physical phenomenon is complex as shown by Bonniot \cite{bonniot2018} with a high degree of uncertainty even considering perfectly known external loads. In practice, dynamic variation, residual and thermal stresses, and natural wear make predicting the phenomenon even more challenging.\\
Data-based modeling offers a viable solution through ground truth observations. However, the difficulty of collecting large-scale data and dealing with poor-quality ultrasonic measurements has resulted in a scarcity of studies in the literature.
\subsection{Uncertainty Estimation.}
Uncertainty quantification (UQ) has become a crucial aspect in various fields, ensuring robust and reliable decision-making in the face of uncertain conditions.\\
Bayesian models for NNs introduce a probabilistic perspective by incorporating prior distributions on the model parameters. This enables a more nuanced estimation of uncertainty, extending beyond point estimates. Recently, Gal and Ghahramani \cite{gal2016dropout} proposed using Monte Carlo dropout (MC-dropout) to estimate
predictive uncertainty by applying Dropout \cite{srivastava2014dropout} at test time. Kendall et al. \cite{kendall2017uncertainties} also suggested an efficient way to capture both aleatoric and epistemic uncertainties within the same model. \\
Ensemble methods involve training and combining multiple neural networks (NNs) models to improve the robustness of predictions. Lakshminarayanan et al. \cite{lakshminarayanan2017simple} introduced a simple and scalable method for predictive uncertainty estimation using deep ensembles. This approach leverages the diversity of individual models to provide a more comprehensive view of uncertainty.\\
Deep evidential networks represent a recent paradigm where the model directly outputs a distribution over predictions, providing a clear characterization of uncertainty. Sensoy et al. \cite{sensoy2018evidential} introduced evidential deep learning to quantify classification uncertainty. This approach explicitly models the epistemic uncertainty associated with each prediction.\\
In this study, the Bayesian approach was employed as it offers a robust framework for uncertainty quantification and predictive modeling in rail crack propagation forecasting. It extends traditional neural networks by incorporating Bayesian inference, thereby transforming the model output into a probability distribution rather than a single point estimate. This probabilistic perspective allows to effectively capture both aleatoric (data-related) and epistemic (model-related) uncertainties, which are critical in scenarios with sparse or noisy data.
Compared to ensemble methods for example, which require training and combining multiple models, Bayesian Neural Networks (BNNs) simplify the architecture by integrating uncertainty estimation within a single model. This approach reduces computational complexity and facilitates the incorporation of domain knowledge into the modeling process.
\subsection{Learning Under Constraints}
In recent years, the integration of domain knowledge into NNs has gained significant attention for its potential to improve model consistency and generalizability. Techniques such as knowledge integration during training (\cite{nandwani2019primal}; \cite{von2021informed}) have been proposed. These methods involve integrating constraints during the learning phase by defining differentiable functions that express constraint violations. This allows the model to learn and minimize violations within the loss function, enhancing consistency with the given constraints. Constraint integration has been explored in various domains, including biology \cite{pulvermuller2021biological}, text generation \cite{DBLP:journals/corr/abs-2010-12884}, physics (\cite{beucler2021enforcing}; \cite{weber2021constrained}), image processing \cite{oktay2017anatomically}, etc.\\
In addition to constraint-based approaches, Physics-Informed Neural Networks (PINNs) (\cite{karniadakis2019physics};\cite{raissi2019physics}) have gained prominence. PINNs leverage physical principles and partial differential equations to guide the training process. This approach facilitates the incorporation of domain-specific knowledge into the neural network architecture.
This involves applying constraints using innovative techniques such as personalized loss functions, adversarial training and hybrid models that combine traditional physics-based methodologies with neural network architectures.\\
Our approach integrates constraints directly into the loss function, so that their impact on model accuracy can be parameterized. This allows to balance model precision with respect for specific physical or domain constraints using adjustable coefficients. Unlike PINNs, which solve differential equations within the network, our method offers flexibility by adapting to the specific requirements of the problem without strictly adhering to predefined physical equations.
\\ \\
In this paper, we present a new contribution to the prediction of crack propagation in rails  by implementing a recurrent Bayesian model capable of considering multiple explanatory variables, quantifying uncertainties and incorporating physical constraints. By integrating domain knowledge and physical principles into the model, predictive power and reliability are enhanced. This approach not only provides a better understanding of crack propagation, but also promotes more effective maintenance strategies.
\section{Methodology}
\label{sec:methodo}
We propose in this paper a Recurrent Bayesian Multi-Horizon model (B-MH) with constraints representing a significant departure from conventional approaches. Our methodology introduces these constraints to ensure the adherence of predictions to specific criteria, thereby strengthening the reliability of the model.
\\ \\
\textbf{Notation}\\
In this study, we conduct a comprehensive analysis of $N$ distinct multivariate time series, represented as pairs $\{(X_i, Y_i)\}_{i=1}^{N}$. Each time series pair is observed over a common temporal domain discretized into $T$ regularly spaced points. 
The dynamic behavior of these time series is influenced by a set of exogenous factors, denoted by $X_{i}$, where $X_{ij} \in \mathbb{R}^M$, with $M$ the number of exogenous features. The target variable is a time series denoted by $Y_{i}$, where $y_{ij} \in \mathbb{R}$. Here, $i$ indexes the individual time series pairs, ranging from $1$ to $N$. Additionally, within each pair, $j$ indexes the time steps, ranging from $1$ to $T$.\\
The inclusion of exogenous variables enriches the modeling approach, enabling the capture of external influences on time series dynamics.\\
For each time series pair $(X_i, Y_i)$, two distinct time horizons are considered: a past horizon of length $t$ and a prediction horizon of length $k$, such that $t + k = T$. The past observations and their corresponding exogenous factors are denoted as $(X_{i,1:t}, Y_{i,1:t})$, and the future predictions and their corresponding exogenous factors are denoted as $(X_{i,t+1:t+k}, Y_{i,t+1:t+k})$.

\subsection{Bayesian Multi-Horizons model}
This model learns to predict the future values of the time series $Y_{i,t+1:t+k}$ considering both historical context $ X_{i,1:t}$ and lengths $Y_{i,1:t}$, as well as the current context $X_{i,t+1:t+k}$. The aim is to model the distribution:
\begin{equation}
    P(Y_{i,t+1:t+k}|Y_{i,1:t}, X_{i,1:t},X_{i,t+1:t+k}).
\end{equation} 
In the context of crack length measurements, it is crucial to identify and address two main forms of uncertainty. The first, known as aleatoric uncertainty, is associated with the inherent noise in observations. This type of uncertainty, representing variability that cannot be reduced by acquiring more data but can be quantified, is always present in crack length measurements. The second type, known as epistemic uncertainty, arises from the uncertainty in model predictions due to insufficient training data for some length measurements. This type of uncertainty can be mitigated by gathering additional data.\\
To incorporate uncertainty estimation, we adapted this model using a Bayesian approach proposed by Kendall et al. \cite{kendall2017uncertainties}, resulting in the Bayesian Multi-horizons model (B-MH). The model's output includes both a predictive mean ($\hat{y}_{ij}$) and predictive variance ($\hat{\sigma}_{ij}^2$), where $\hat{y}_{ij}$ represents the predictive mean of the crack length, and $\hat{\sigma}_{ij}^2$ represents its predictive variance. The general architecture of the model is described in figure \ref{fig1}.\\
To model random uncertainty, a Gaussian likelihood is employed, conforming to the distribution characteristics of the available crack length values. The loss function of the B-MH model is formulated as a combination of the mean square error (MSE) loss and a regularization term, designed to make the model robust to noisy data while preventing predictions of infinite uncertainty. For numerical stability, the predictive variance term is replaced by $s_{ij} = \log (\hat{ \sigma}(X_{ij})^2)$.\\
The weights of the MSE and regularization terms are set to $\frac{2}{3}$ and $\frac{1}{3}$, respectively, prioritizing the minimization of MSE over regularization. The final minimization function is expressed as:
\begin{equation}
 L_\mathrm{B\_MH}\ =\ \frac{1}{N*T} \sum_{i=1}^{N} \sum_{j=1}^{T} \frac{2}{3} exp(- s_{ij}) ||y_{ij}-\hat{y}_{ij}||^2 +  \frac{1}{3} s_{ij}.
\end{equation}
To quantify uncertainty, a dropout approach \cite{srivastava2014dropout} is employed during training and inference, generating stochastic predictions. Predictive uncertainty is approximated using $S$ stochastic prediction samples, differentiating between epistemic and aleatoric uncertainties. The predictive total uncertainty for a time step $i$ of a time series $j$ can be approximated using
\begin{equation}
\mathrm{Var}(y_{ij})\ \approx\ \left(\frac{1}{S}\sum_{s=1}^{S} \hat{y}_{ij(s)}^2 - \left(\frac{1} {S}\sum_{s=1}^{S} \hat{y}_{ij(s)}\right)^2 \right) + \frac{1}{S} \sum_{s=1}^{S} \hat{\sigma}_{ij(s)}^2,
\label{eq6}
\end{equation}
with $\{{ \hat{y}_{ij(s)}, \hat{\sigma}_{ij(s)}^2 }\}_{s=1}^{S}$ the set of $S$ sampled outputs after each forward pass.\\
The first term of this total variance corresponds to the epistemic uncertainty and the second one corresponds to the aleatoric uncertainty.
\begin{figure*}[htbp]
    \centering
    \includegraphics[scale=0.75]{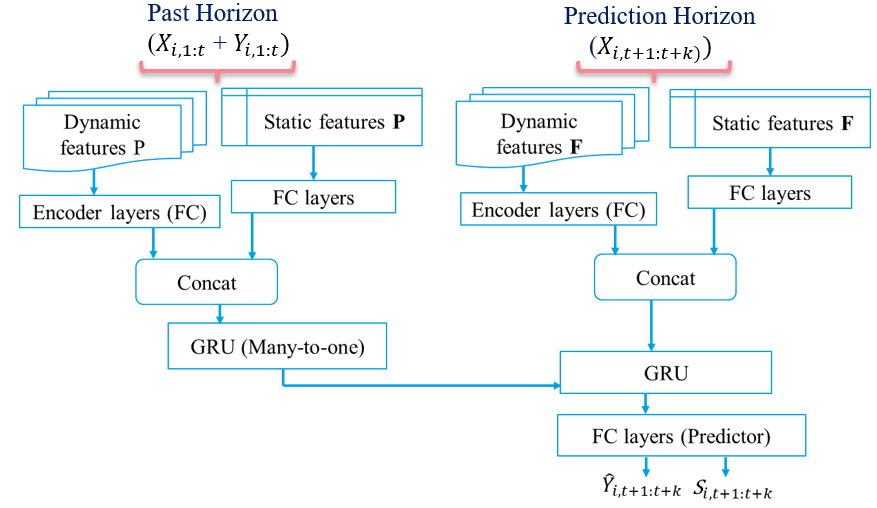}
    \caption{General architecture of the Baysian Multi-Horizons model (B-MH)}
    \label{fig1}
\end{figure*}
\subsection{Constraints integration}
\subsubsection{Monotonicity Constraint}
Crack can not physically close but some external event such as compression due to high temperature can induce drops in crack length measured by the ultrasonic measurement system as explained in \cite{kou2021fully}, thus inducing a measurement error. These drops are kept in data but should not be reproduced by the model. Therefore, to guarantee monotonic growth of predicted values, a fundamental criterion for accurate rail crack length predictions, we introduce a monotonicity constraint. This constraint, integrated into the loss function, is designed to ensure conformity with monotonic trends. Mathematically, the monotonicity constraint is expressed for each time step $j$ of a time series $i$ as:
\begin{equation}
L_\mathrm{monotonicity,_{ij}} = \max(0, \hat{y}_{ij-1} - \hat{y}_{ij}),
\end{equation}
where $\hat{y}_{ij}$ is the predicted mean at time step $j$ of a time series $i$.
\subsubsection{Asymmetry Constraint}
To ensure that the model does not underestimate crack lengths, particularly when they become important, an asymmetry constraint is introduced. This constraint is designed to improve compliance with the nuanced dynamics of large values and is expressed mathematically as: 
\begin{equation}
L_\mathrm{asymmetry,_{ij}} = \max(0, y_{ij} - \hat{y}_{ij}).
\end{equation}\\
    This constraint is multiplied by a factor, arbitrary set at $log(2+y)$. $Scale$ is a boolean parameter that controls whether this multiplication is performed or not. This factor dynamically scales the constraint based on the values of $y$, introducing increased penalization as $y$ values become critical. The addition of $2$ to $y$ prevents multiplication by zero in cases where $y$ is zero. Thus, the constraint becomes : \\
\begin{equation}
\begin{split}
L_\mathrm{asymmetry,_{ij}} = \left\{
    \begin{array}{ll}
        \max(0, y_{ij} - \hat{y}_{ij}) \times log(2+y_{ij}) \\ \hfill \hfill \hfill \hfill \hfill \hfill \hfill \mbox{if} \ scale = true \\ \\
        \max(0, y_{ij} - \hat{y}_{ij}) \\ \hfill \hfill \hfill  \mbox{otherwise.}
    \end{array}
\right.
\end{split}
\end{equation}

\subsection{Total New Loss Function}
The way to introduce these constraints into the model represents a new challenge. This can be done simply by summing the Bayesian cost function $L_\mathrm{B\_MH}$ with these 2 constraints, or by introducing these two constraints within the Bayesian loss.
\subsubsection{Losses Summation}
The first way of integrating these constraints into the total cost function is 
defined as the sum of three terms, each with its respective weight:
\begin{equation}
\begin{split}
L_{total, sum} =  L_\mathrm{B\_MH} + \frac{1}{N*T} \sum_{i=1}^{N} \sum_{j=2}^{T} (\beta  L_\mathrm{monotonicity,_{ij}} \\+ \lambda L_\mathrm{asymmetry,_{ij}}),
\end{split}
\end{equation}
where $\beta$ and $\lambda$are weights that control the influence of the monotonicity and asymmetry constraints, respectively.

\subsubsection{Bayesian Loss Alteration}
Another way for incorporating these constraints is to modify the Bayesian cost function itself. The total cost function is then expressed as:
\begin{equation}
\begin{split}
 L_\mathrm{total, bayes}\ = \frac{1}{N*T} \sum_{i=1}^{N} \sum_{j=1}^{T} \frac{2}{3} exp(- s_{ij}) [~||y_{ij}-\hat{y}_{ij}||^2 \\ + \beta  L_\mathrm{monotonicity,_{ij}} + \lambda L_\mathrm{asymmetry,_{ij}}~] +  \frac{1}{3} s_{ij}.
 \end{split}
\end{equation}
This modified total cost function represents an alternative approach, integrating constraints directly into the Bayesian model cost function, offering a comprehensive strategy for optimizing the B-MH model while taking into account both uncertainty modeling and specified constraints.
\section{Experiments}
\label{sec:exp}
\subsection{Data}
\label{data}
The dataset used in this work includes various defects in the rail network, each with multiple records of crack lengths observed over several visits. The temporal aspect is highlighted by the dates of these visits, transforming the dataset into time series. Four main categories characterize the data:
\begin{itemize}
    \item \textbf{Infrastructure Data} Detailing railway network features influencing vehicle dynamics, including rail linear mass, sleeper type, rail grade, and more.
    \item \textbf{Traffic Data} Capturing the dynamic impact of rolling stock through variables such as maximal velocity, acceleration, braking, annual tonnage, and vehicle types.
    \item \textbf{Environment Data} Incorporating non-railway environmental factors, such as temperature and rain, categorized by intensity.
    \item \textbf{Defect Data} This variable is of particular interest as it serves as the target for prediction. It focuses on specific rail defects, such as squats, which are recorded on various dates and regularly inspected to monitor the evolution of crack lengths.
\end{itemize}
While the dataset provides a rich source of information, with "Defect Data" as the target variable, complexities arise in the crack data, including anomalies in length values, diverse discovery dates, and varying intervals between visits. Preprocessing was essential to ensure dataset consistency.\\
All information from the various datasets was consolidated into a single training dataset. Anomalies identified in the data were addressed based on expert knowledge. To manage irregular time steps in the time series, interpolation was conducted, with a chosen frequency of 3 months, as described in \cite{ouerk2023rail}.\\
After this step, defects exhibiting a decrease in values greater than 15mm were removed from the database to prevent errors in the learning model. However, drops of less than 15mm were tolerated, as they could be attributed to variations in measurement conditions, such as temperature changes leading to crack closure. Additionally, the measurement process is subject to operator interpretation, introducing variability between operators. 
This process resulted in approximately 60000 time series, each having a 3-month time-step frequency and a maximum length of 59 time steps.
\subsection{Settings}
The work has been implemented in Python using Pytorch library. All the experiments are conducted using an  NVIDIA A40 GPU.\\
Adam optimizer is used to perform the gradient descent minimization of the loss function. The activation function used is the $Tanh$ function for all hidden layers.\\
To benchmark the different models, we considered a multi-criteria approach with several categories of criteria : ML, physical compliance, and industrial readiness as suggested in \cite{leyli2022lips}, \cite{yagoubi2024neurips}. We used  MAE (Mean Absolute Error) and RMSE (Root Mean Square Error)  as ML metrics. Some physical criteria are considered to control the physical constraint violation consisting in the drop in the crack length. 
\begin{itemize}
\item{\textbf{MSQNS}}, for Mean SeQuence Negative Slope, is the percentage of sequences that contain at least one fall in the predicted values;
\item{\textbf{MSTNS}}, for Mean STeps Negative Slope, is the percentage of steps that contain at least one fall in the predicted values. 
\item{\textbf{MLNS}}, for Mean Length Negative Slope, is the mean value of the fall in predicted length values. As a reminder, the observation time series themselves contain drops in values of up to 15mm.
\end{itemize}

Other metrics are used to show the effect of the asymmetry constraint. These criteria are:
\begin{itemize}
\item{\bm{$\% \mathbb{1}[\hat{y} < y]$}}, for the percentage of samples in which the prediction is lower than the actual value. 
\item{\bm{$\% \mathbb{1}[\hat{y} < y]_{80}$}}, for the percentage of samples in which the prediction is lower than the actual value when this actual value is $ \geq 80mm$. 
\end{itemize}
For all the experiments, time series were created using a sliding window of size 9: with a past horizon of size 5 and a prediction horizon of size 4. The size of the past horizon containing historical crack values was chosen at 5 time steps, inspired by \cite{lara2021experimental} which suggests that a past horizon of size $1.25 \times k$ ($k$ being the size of the prediction horizon) gives the best prediction results. After choosing this maximum size for the prediction horizon, not all the series generated have the same length. As some are shorter than the maximum length, these series have been completed by adding zeros at the end, so that they all have the same length. These completed time steps will be ignored when calculating the cost functions by implementing custom functions that ignore these time steps for backpropagation. The dataset was divided into three parts: 60 \%  training set for the learning procedure, 20 \% validation set for hyperparameter optimization and convergence control, and 20 \% test set for performance evaluation.
The division strategy adopted ensures that the subsequences of a given defect series belong to only one of the three previous sets.\\
The time series are then normalized using a custom time series standard scaler, so that their mean is 0 and their standard deviation is 1. This makes the model much more robust to outliers. Min-max normalization has also been tested, but gives slightly poorer results.\\All models were trained with a dropout rate of $10\%$. The B-MH model converges after 10 epochs. At inference time, for this model (B-MH), in order to quantify uncertainty, $50$ stochastic outputs were predicted for each sample ($S = 50$). 
\subsection{Comparison of the B-MH model to Baselines}
Our Bayesian Multi-Horizon (B-MH) model (before integrating constraints) has been compared to several benchmark models in order to evaluate its performance. These models are all based on recurrent neural networks, and are briefly described as follows:
\begin{itemize}
    \item \textbf{RNN-FC, LSTM-FC and GRU-FC: } are simply RNN, LSTM, and GRU models, respectively, augmented with Fully Connected layers to encode static features. These models take into account all exogenous variables but not the historical crack length values.
    \item \textbf{RNN-FC-LH, LSTM-FC-LH and GRU-FC-LH: } are the same as the previous models but incorporate historical crack length values.
    \item \textbf{Multi-Horizons: } This model is a recurrent neural network with multiple time horizons. It consists of a past horizon which takes as input exogenous variables and historical crack length measurements, and a future prediction horizon which takes as input the encoded output from the past horizon as well as current contextual variables in order to infer future crack length values
\end{itemize}
Table \ref{tab1} illustrates the ML performance of these different models in terms of MAE and RMSE in comparison with the Bayesian Multi-Horizon (B-MH) model. As evident, this model not only quantifies uncertainties but also surpasses these baseline models in ML score performance.
\begin{table}[]
\centering
\caption{MAE and RMSE scores for the first time step in the prediction horizon, and mean MAE and RMSE over the prediction horizons of the B-MH model and the baselines}
\begin{tabular}{l|l|l|l|l}
\hline
Mode                                  & \cellcolor[HTML]{EFEFEF}\begin{tabular}[c]{@{}l@{}}MAE \\ 1st\end{tabular} & \cellcolor[HTML]{EFEFEF}\begin{tabular}[c]{@{}l@{}}Mean \\ MAE\end{tabular} & \cellcolor[HTML]{EFEFEF}\begin{tabular}[c]{@{}l@{}}RMSE\\ 1st\end{tabular} & \cellcolor[HTML]{EFEFEF}\begin{tabular}[c]{@{}l@{}}Mean \\ RMSE\end{tabular} \\ \hline
\cellcolor[HTML]{C0C0C0}RNN-FC         & 10.48                                                                      & 10.47                                                                       & 13.67                                                                      & 13.66                                                                        \\ \hline
\cellcolor[HTML]{C0C0C0}LSTM-FC        & 10.54                                                                      & 10.53                                                                       & 13.75                                                                      & 13.72                                                                        \\ \hline
\cellcolor[HTML]{C0C0C0}GRU-FC         & 9.65                                                                       & 9.45                                                                        & 12.60                                                                      & 12.38                                                                        \\ \hline
\cellcolor[HTML]{C0C0C0}LSTM-FC-LH     & 2.37                                                                       & 3.45                                                                        & 4.72                                                                       & 6.01                                                                         \\ \hline
\cellcolor[HTML]{C0C0C0}GRU-FC-LH      & 2.37                                                                       & 3.49                                                                        & 4.77                                                                       & 6.06                                                                         \\ \hline
\cellcolor[HTML]{C0C0C0}Multi-Horizons & 1.54                                                                       & 2.64                                                                        & 2.62                                                                       & 4.33                                                                         \\ \hline
\cellcolor[HTML]{C0C0C0}B-MH           & \textbf{0.94}                                                              & \textbf{2.19}                                                               & \textbf{2.37}                                                              & \textbf{4.06}                                                                \\ \hline
\end{tabular}
\label{tab1}
\end{table}

\subsection{Constraint parameter Optimization}
To enhance our model's performance and ensure adherence to constraints, we employed the widely-used multiobjective evolutionary algorithm NSGA-II \cite{deb2002fast} to simultaneously optimize the parameters. This optimization process aimed to fine-tune the coefficients associated with asymmetry ($\lambda$) and monotonicity ($\beta$) to enhance model behavior. The optimization objectives encompassed three metrics: MAE, MSTNS and $\% \mathbb{1}[\hat{y} < y]$. This optimization was conducted for both types of constraints integration—Losses Summation and Bayesian Loss Alteration—using a budget of 200 trials. For implementing the multiobjective hyperparameters optimization, we utilized the Optuna framework \cite{akiba2019optuna}. The $\textbf{ReLU}$ function was used to implement the two constraints, in order to ensure their differentiabily.  \\
Upon completion of the multi-objective hyperparameters optimization, we relied on the Pareto diagram to visually capture the solutions representing the optimal compromise, revealing the intricate relationship between the optimized parameters and the model's objectives. Furthermore, we examined the correlation matrix of the coefficients obtained, which provided valuable insights into the relationships between hyperparameters and model constraints.\\
Negative correlations between a coefficient and a metric indicate that an increase in the coefficient corresponds to a decrease in the corresponding metric, which is favorable in our goal to minimize MSTNS and $\% \mathbb{1}[\hat{y} < y]$.\\ 

\noindent \textbf{Losses Summation}\\
Figure \ref{fig:pareto} illustrates the trade-off between compliance with constraints (reduction of MSTNS and $\% \mathbb{1}[\hat{y} < y]$) and model precision (MAE). The plotted points are taken from the Pareto front of this multi-objective optimization. Points along the Pareto front represent efficient solutions where improvements in one criterion come at the expense of another. Each point reflects a unique combination of $\lambda$ and $\beta$, highlighting the delicate balance required for model optimization.\\
As observed from this figure, there exists an inverse relationship between MAE and the $\% \mathbb{1}[\hat{y} < y]$ metric. Indeed, a reduction in one metric appears to have an adverse impact on the other (increase). Conversely, reducing the MSTNS metric does not seem to notably affect MAE.

\begin{figure*}[!t]
    \centering
    \subfloat[MAE vs MSTNS]{%
        \includegraphics[width=0.45\textwidth]{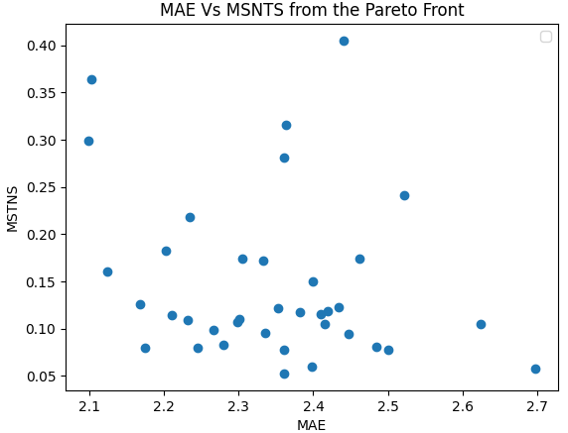}
    }
    \hfill
    \subfloat[MAE Vs ${\% \mathbb{1}[\hat{y} < y]}$ ]{%
        \includegraphics[width=0.45\textwidth]{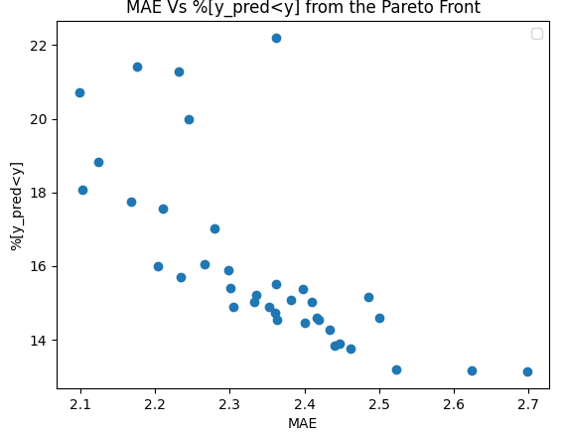}
    }
    \caption{Pareto front of multi-objective hyperparameter optimization with Optuna}
    \label{fig:pareto}
\end{figure*}

\begin{figure}[h!]
    \centering
    \includegraphics[scale=0.75]{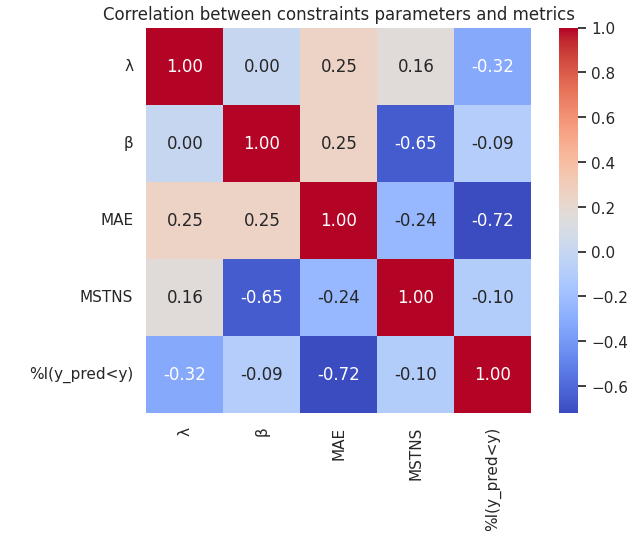}
    \caption{Correlation between the constraints parameters ($\lambda$ and $\beta$) and metrics for Losses Summation}
    \label{fig3}
\end{figure}
\noindent Figure \ref{fig3} shows the correlation results between the constraints parameters and the different metrics. \\
\indent \textbf{Parameter} \bm{$\lambda$}: A positive correlation between $\lambda$ and MAE suggests that an increase in $\lambda$ corresponds to a rise in MAE. This implies that the regularization effect introduced by $\lambda$  may lead to a slightly higher MAE. On the other hand, the negative correlation between $\lambda$ and $\% \mathbb{1}[\hat{y} < y]$ shows that higher $\lambda$ values help the model to better satisfy the asymmetry constraint. The positive correlation with MSTNS indicates that an increase in $\lambda$ is associated with a modest increase in the MSTNS score.\\
\indent \textbf{Parameter} \bm{$\beta$}: The positive correlation with the MAE suggests that an increase in $\beta$ is associated with a higher mean absolute error. This implies that higher weighting on the monotonicity constraint could lead to a slight sacrifice in predictive accuracy. The negative correlation with $\% \mathbb{1}[\hat{y} < y]$ indicates a potential, though weaker, impact of $\beta$ on the percentage of predictions below real values. This suggests that the monotonicity constraint may have a subtle influence on the asymmetry constraint. The strong negative correlation with $MSTNS$ $(-0.65)$ highlights that an increase in $\beta$ contributes significantly to reducing the $MSTNS$ metric. This is in line with the objective of minimizing this score.\\ \\
\textbf{Bayesian Loss Alteration} 
\begin{figure}[h!]
    \centering
    \includegraphics[scale=0.75]{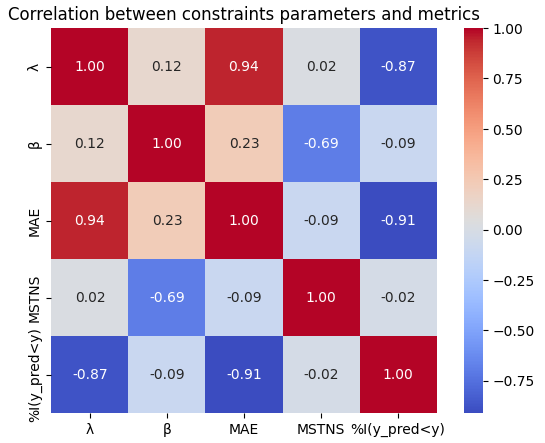}
    \caption{Correlation between the constraints parameters ($\lambda$ and $\beta$) and metrics for Bayesian Loss Alteration}
    \label{fig4}
\end{figure}

\indent \textbf{Parameter} \bm{$\lambda$}:
\begin{itemize}
  \item MAE: $\lambda$ has a strong positive correlation (0.9416) with MAE. As $\lambda$ increases, MAE tends to increase significantly.
  \item MSTNS: $\lambda$ has a weak positive correlation (0.0169) with MSTNS. The impact of $\lambda$ on MSTNS is minimal.
  \item $\% \mathbb{1}[\hat{y} < y]$: $\lambda$ has a strong negative correlation (-0.8744) with $\% \mathbb{1}[\hat{y} < y]$. An increase in $\lambda$ corresponds to a significant decrease in $\% \mathbb{1}[\hat{y} < y]$.
\end{itemize}

\indent \textbf{Parameter} \bm{$\beta$}:
\begin{itemize}
  \item MAE: $\beta$ has a negligible impact on MAE, with a weak positive correlation (0.1221).
  \item MSTNS: $\beta$ has a moderate negative correlation (-0.6910) with MSTNS. An increase in $\beta$ tends to decrease MSTNS.
  \item $\% \mathbb{1}[\hat{y} < y]$: $\beta$ has a negligible impact on $\% \mathbb{1}[\hat{y} < y]$, with a weak negative correlation (-0.0926).
\end{itemize}
In summary, $\lambda$ has a high impact on MAE and $\% \mathbb{1}[\hat{y} < y]$, with an increase in $\lambda$ leading to higher MAE and lower $\% \mathbb{1}[\hat{y} < y]$. On the other hand, $\beta$ has a noticeable impact on MSTNS, where an increase in $\beta$ leads to a decrease in MSTNS. Correlations provide valuable insights into how adjustments in $\lambda$ and $\beta$ affect the model's performance metrics.
\subsection{Results}
Table \ref{tab2} shows the performance of the constrained model with the $1^{st}$ way of integrating constraints (Loss Summation). The parameters $\lambda$ and $\beta$ chosen for this benchmarking are taken from the Pareto Front at the end of the hyperparameter optimization. Obtained results show that integrating the constraints with Bayesian loss summation reduces the 1st constraint linked to value drops in predictions while preserving a good prediction score (MAE), although the metrics linked to second constraint of asymmetry are not truly reduced.
\begin{table*}[]
\centering
\caption{ML and physical metrics results according to the parameters $\lambda$ and $\beta$ and scale the Losses Summation integration}
\begin{tabular}{c|c|c|c|c|c|c|c|c}
\hline
\rowcolor[HTML]{EFEFEF} 
 $\lambda$ &  $\beta$ & Scale & MAE  & MLNS & MSTNS  & MSQNS  & $\% \mathbb{1}[\hat{y} < y]$ & $\% \mathbb{1}[\hat{y } < y]_{80}$ \\ \hline
 0                                                 & 0                                              & False & 2.19 & 1.07 & 0.8\%  & 1.30\% & 19.5\%                                                                                                                              & 24\%                                                                                                                                    \\ \hline
0.10                                                    & 4                                                    & True                                               & 2.27                                             & 1.03                                              & 0.45\%                                             & 1.73\%                                             & 19.2\%                                         & 25\%                                            \\ \hline
0.10                                                    & 5                                                    & True                                               & 2.21                                             & 1.05                                              & 0.6\%                                              & 2.4\%                                              & 18.4\%                                         & 22.3\%                                          \\ \hline
0.10                                                    & 7                                                    & True                                               & 2.22                                             & 1.02                                              & 0.31\%                                             & 1.21\%                                             & 18.9\%                                         & 22.3\%                                          \\ \hline
1                                                       & 4                                                    & True                                               & 2.40                                             & 1.02                                              & 0.44\%                                             & 1.71\%                                             & 16.5\%                                         & 23.9\%                                          \\ \hline
1.48                                                    & 7.26                                                 & True                                               & 2.33                                             & 1.01                                              & 0.33\%                                             & 1.30\%                                             & 16.2\%                                         & 23.2\%                                          \\ \hline
1.80                                                    & 4.73                                                 & True                                               & 2.38                                             & 1.01                                              & 0.17\%                                             & 0.67\%                                             & 17.00\%                                        & 24.49\%                                         \\ \hline
0.91                                                    & 8.67                                                 & True                                               & 2.43                                             & 1.02                                              & 0.14\%                                             & 0.57\%                                             & 16.98\%                                        & 23.34\%                                         \\ \hline
0.89                                                    & 3.34                                                 & True                                               & 2.32                                             & 1.01                                              & 0.14\%                                             & 0.59\%                                             & 20.63\%                                        & 22.47\%                                         \\ \hline
0.94                                                    & 8.76                                                 & True                                               & 2.42                                             & 1.01                                              & 0.10\%                                             & 0.41\%                                             & 15.04\%                                        & 25.28\%                                         \\ \hline
8.28                                                    & 9.26                                                 & True                                               & 2.40                                             & 1.01                                              & 0.25\%                                             & 1.01\%                                             & 16.54\%                                        & 24.19\%                                         \\ \hline
\end{tabular}
\label{tab2}
\end{table*}

\begin{table*}[h!]
\centering
\caption{ML and physical metrics results according to the parameters $\lambda$, $\beta$ and scale for the Bayesian Loss Alteration integration}
\begin{tabular}{c|c|c|c|c|c|c|c|c}
\hline
\rowcolor[HTML]{EFEFEF} 
 $\lambda$ &  $\beta$ & Scale & MAE  & MLNS & MSTNS  & MSQNS  & $\% \mathbb{1}[\hat{y} < y]$ & $\% \mathbb{1}[\hat{y } < y]_{80}$ \\ \hline
0                                                 & 0                                              & False & 2.19 & 1.07 & 0.8\%  & 1.30\% & 19.5\%                                                                                                                              & 24\%                                                                                                                                    \\ \hline
0.06                                              & 0.26                                           & False & 2.97 & 1.36 & 0.00\% & 0.02\% & 12\%                                                                                                                                & 25\%                                                                                                                                    \\ \hline
0.06                                              & 0.26                                           & True  & 2.75 & 1.00 & 0.01\% & 0.02\% & 11.5\%                                                                                                                              & 17\%                                                                                                                                    \\ \hline
0.06                                              & 0.5                                            & False & 3.08 & 1.33 & 0.00\% & 0.00\% & 14\%                                                                                                                                & 17\%                                                                                                                                    \\ \hline
0.06                                              & 0.5                                            & True  & 2.89 & 1.33 & 0.00\% & 0.00\% & 14\%                                                                                                                                & 17\%                                                                                                                                    \\ \hline
0.05                                              & 0.169                                          & False & 2.97 & 1.12 & 0.02\% & 0.06\% & 11.8\%                                                                                                                              & 17.9\%                                                                                                                                  \\ \hline
0.05                                              & 0.169                                          & True  & 2.62 & 1.28 & 0.01\% & 0.03\% & 13.8\%                                                                                                                              & 19\%                                                                                                                                    \\ \hline
0.33                                              & 0.32                                           & False & 3.92 & 1.24 & 0.02\% & 0.07\% & 8\%                                                                                                                                 & 20\%                                                                                                                                    \\ \hline
0.33                                              & 0.32                                           & True  & 3.92 & 1.09 & 0.02\% & 0.07\% & 8\%                                                                                                                                 & 15.9\%                                                                                                                                  \\ \hline
\end{tabular}
\label{tab3}
\end{table*}
\noindent Table \ref{tab3} shows the performance of the constrained model with the $2^{nd}$ method of integrating constraints (Bayesian Loss Alteration). The $\lambda$ and $\beta$ parameters chosen for this comparative analysis are also extracted from the Pareto front at the end of the hyperparameter optimization. The impact of the scale parameter (mentioned in equation \ref{eq6}) was also checked, turning it on and off each time. Results show that integrating the constraints with Bayesian loss modification significantly reduces the percentage of value drops in predictions, while preserving a good prediction score (MAE). Metrics related to the second asymmetry constraint are also reduced. In fact, by integrating constraints into the inner sum, their impact is multiplied by the uncertainty associated with the data $(exp(s_{ij}))$ in the same way as the squared error, making it easier to manage noisy data. Thus, the inclusion of noise into constraints seems to have an impact on their effectiveness. We thus need to choose parameters $\lambda$ and $\beta$ that satisfy the constraints while preserving acceptable precision (MAE). The triplet of parameters ($0.06, 0.26, True$) seems to give a good overall result.\\

\section*{Conclusion}
\label{sec:conclusion}

In this study, we introduced a constrained Multi-Horizons Bayesian RNN which allows the consideration of physical and risk aspects of a rail crack propagation. Obtained results have shown that considering constraints inside the Bayesian loss has a relevant impact on the global performance of the proposed architecture, with a significant improvement with regard to physical and risk performance indicator. A multiobjective hyperparameters tuning approach was conducted to improve the trade-off between the considered criteria. In addition, we propose a Bayesian framework to help quantifying uncertainties, and providing indications of the predictions reliability. 
The contextualization in the industrial environment has shown that the trade-off linked with the dropout ratio was not relevant. Finally, by simultaneously accounting for constraints and uncertainties, our model offers a more robust assessment of rail crack propagation forecasting, enabling decision-makers to adopt efficient maintenance strategies and minimize the risk of unexpected failures.
While our study focused on RNN-based architectures, primarily due to the relatively short length of the time series data, which RNNs handle efficiently, we acknowledge the potential of transformer-based models for future work. Transformers \cite{vaswani2017attention}, with their ability to handle long-range dependencies and parallelize computations, could be beneficial in scenarios involving longer sequences or more complex dependencies.
Future research will explore the application of transformer models with built-in mechanisms for incorporating constraints. This could offer a more advanced predictive modeling that may optimize maintenance strategies in railway infrastructure management.

\bibliographystyle{IEEEtran}
\bibliography{bibliography}

\end{document}